# Knowledge Retrieval Using Functional Object-Oriented Networks

Gabriel Laverghetta

*Abstract*—Robotic agents often perform tasks that transform sets of input objects into output objects through functional motions. This work describes the FOON knowledge representation model for robotic tasks. We define the structure and key components of FOON and describe the process we followed to create our universal FOON dataset. The paper describes various search algorithms and heuristic functions we used to search for objects within the FOON. We performed multiple searches on our universal FOON using these algorithms and discussed the effectiveness of each algorithm.

## I. Introduction

Many tasks performed by robots are functional in nature. These tasks involve an input object or set of objects that is transformed into an output object through a motion. The output produced by the task depends upon the performed motion. To complete these tasks, robots must be able to assess the objects involved and the motions taking place. This emphasis on both objects and motions also occurs in the human brain [1].

Developing a robust form of knowledge representation is important for robots to complete their tasks efficiently and successfully. A Functional Object-Oriented Network (FOON) is a knowledge representation designed for these tasks. FOONs contain various input and output object nodes connected to each other through motion nodes. FOONs model the relationships among objects and the changes that objects undergo due to motions. They may also contain information related to object state such as temperature, ingredients, etc. The FOON representation has proven to be particularly useful for robotic cooking [2].

A single FOON graph will likely contain far more information than the robot will need to complete a given task. For example, a FOON may contain hundreds of recipes along with the ingredients and steps needed to prepare these recipes. With such a vast quantity of information available, it is important to employ algorithms that quickly search the FOON and retrieve the sequence of steps needed to prepare the desired recipe. This work describes three search algorithms that have been developed for FOON and discusses their effectiveness.

## II. Video Annotation and FOON creation

First, we review the basic FOON terminology and our video annotation procedure. We used this procedure to generate our universal FOON, which served as the dataset for our experiments.

A FOON consists of interconnected *functional units,* each representing a single action. An example of a functional unit is shown below:

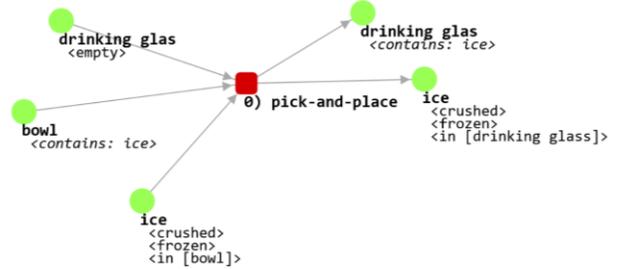

Fig. 1. A sample functional unit.

Arrows extend outward from the input object nodes into the motion node. From the motion node, arrows extend into the output object nodes. Object states are also shown below the object names. Note that not all input objects may be changed by a motion; in the example functional unit, the bowl object still contains some ice after the motion takes place.

A *subgraph* is formed by chaining multiple functional units together. Subgraphs are used to represent a single, complete activity, such as a cooking recipe. An example of a subgraph for a macaroni recipe is as follows:

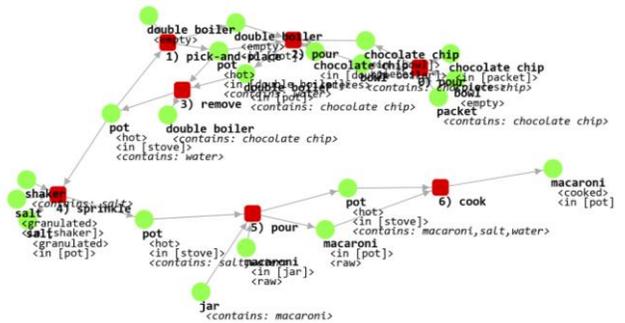

Fig 2: Example subgraph for a macaroni recipe. It contains seven functional units.

In addition to the graphical display shown above, subgraphs may also be described in text files using specific syntax. For this paper, we created the text files manually. Using cooking demonstration videos as a reference, we annotated the videos and wrote functional units for the motions shown in the videos.

Subgraphs may be combined to form a *universal FOON*. When merging subgraphs into a universal FOON, it is important to account for duplicate functional units. Two functional units are considered duplicates if they have the same input and output object sets and the same type of motion node. The merging algorithm used to create universal FOONs avoids creating duplicate functional units [2]. After we finished the video annotation process, each of the subgraphs we created was merged to create the universal FOON. This FOON, shown in the figure below, contains 2,376 functional units, and its text file is 35,576 lines long.

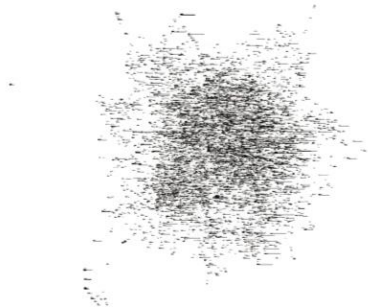

Fig. 3. Zoomed-out graphical view of our universal FOON.

*A task tree* is a sequence of functional units that results in a desired goal node. Task trees are derived by searching a universal FOON for a specific object. The search generates a subgraph showing the functional units involved in creating the goal object node. As with any search problem, there may be numerous paths that lead to the goal node. Our primary purpose in this work is to discuss the effectiveness of several FOON search algorithms.

## III. METHODOLOGY

For the purposes of this work, the universal FOON is organized as a graph, allowing us to make use of graph search algorithms. Our searches always begin at the goal node, which is the desired item we wish to make. The search concludes when each of the leaf nodes in the search tree is available in the kitchen (i.e., ready for use without needing further modification). We tested two different search algorithms: iterative deepening search and greedy best-first search.

*Iterative deepening search* (IDS) repeatedly performs depth-first search within a depth limit. The depth limit begins at zero, allowing only the starting node to be searched. When the depth limit is one, the starting node and its immediate neighbors may be explored. The depth limit continues to increase until the goal is found. If the search fails to find the goal within the depth limit, the search restarts entirely, increasing the limit by one. Using this algorithm combines some of the strengths of depth first search with some of the strengths of breadth first search. Pseudocode for IDS is as follows:

---

**Algorithm III.1** Iterative deepening search

**Input**: s: Search space
g: Goal node
d: Depth bound (initially zero)
**Output**: True if g ∈ s, false otherwise
stack ← []
nodesOutsideLimit ← false
stack.push(goal)
**while** stack is not empty **do**
    currentItem ← stack.pop()
    **if** currentItem = goal **then**
        **return** true
    **end if**
    **for** node **in** currentItem.neighbors **do**
        **if** node.depth <= d **then**
            stack.push(node)
        **else**
            nodesOutsideLimit ← true
        **end if**
    **end for**
**end while**

**if** nodesOutsideLimit = true **then**
    **return** IDS(s, g, d + 1)
**end if**
**return** false

---

*Greedy best-first search* performs breadth-first search guided by a *heuristic function* h(n). When choosing a new node to explore, it searches the available nodes for the node with the highest value of h(n), rather than always choosing the node at the front of the queue. Pseudocode for this algorithm is shown below:

---

**Algorithm III.2** Greedy best-first search

**Input**: s: Search space
g: Goal node
h: heuristic function
**Output**: True if g ∈ s, false otherwise
queue ← []
queue.push_front(goal)
**while** queue is not empty **do**
    currentItem ← queue.pop()
    **if** currentItem = goal **then**
        **return** true
    **end if**
    highest_h_val ← -∞
    **for** node **in** currentItem.neighbors **do**
        **if** h(node) > highest_h_val **then**
            highest_h_val ← h(node)
            next_node ← node
        **end if**
    **end for**
    queue.push_front(next_node)
**end while**
**return** false

The use of a heuristic function adds an informed element to the search. Such heuristics may greatly increase the speed of the search or satisfy other desired parameters. The two heuristic functions we used are as follows:

$$h(n) = \text{motion success rate} \quad (1)$$

$$h(n) = \text{number of input objects} \quad (2)$$

Heuristic (1) considers the *success rate* of the functional unit's motion. This is the probability that the robotic agent will successfully perform the motion. In our case, for example, the robot has a 90% chance of completing pouring motions successfully, but only a 10% chance of completing chopping motions without failing. By employing this heuristic, we can avoid having to perform motions that are unlikely to succeed.

Heuristic (2) considers the size of the functional unit's input object set. Functional units with fewer input objects are favored over those with many inputs. Note that input object ingredients are also included in this heuristic. For example, a bowl containing salt and pepper is regarded as two separate objects. The goal of this heuristic is to prioritize simpler motions with fewer objects involved.

## IV. EXPERIMENT/DISCUSSION

There are several factors we considered when judging the effectiveness of each search algorithm. These include the time and space complexities, the number of functional units in the task trees, and the average success rates of motions performed. Breadth-first search serves as the baseline when evaluating the other algorithms.

The time and space complexities of each algorithm depend upon two values: the *depth* of the solution d and the *branching factor* at each non-leaf node b. The depth is a measure of how deep the goal is located in the search space, and the branching factor refers to the average number of neighbors adjacent to non-leaf nodes. The following table summarizes the time and space complexities of each search algorithm.

Table 1: Time and space complexity

| Algorithm | Time complexity | Space complexity |
|---|---|---|
| BFS | $O(b^d)$ | $O(b^d)$ |
| IDS | $O(b^d)$ | $O(b * d)$ |
| Greedy best-first | $O(b^d)$ | $O(b^d)$ |

In practice, each algorithm has additional advantages and disadvantages. BFS takes longer to find goal nodes located deep within the search space. IDS finds longer paths more quickly, but there is redundant effort in the search due to the depth limit. The performance of Greedy best-first search is affected by the heuristic function; a more complicated function will slow the search.

Aside from reducing time and memory requirements, we also sought to reduce the number of functional units in the generated task trees. With fewer functional units, there are fewer motions that need to be performed to prepare the recipe. To evaluate this factor, we tested each algorithm with several sample goal nodes and measured the number of functional units in the respective task trees. These results are summarized in Table 2.

Table 2: Number of functional units in the task trees

| Goal node | BFS | IDS | GBFS, heuristic (1) | GBFS, heuristic (2) |
|---|---|---|---|---|
| whipped cream | 10 | 10 | 10 | 15 |
| Greek salad | 28 | 28 | 35 | 31 |
| macaroni | 7 | 7 | 7 | 8 |
| sweet potato | 3 | 3 | 3 | 3 |
| ice | 1 | 1 | 1 | 1 |
| quesadilla | 29 | 29 | 29 | 29 |
| sandwich | 13 | 13 | 13 | 13 |

These results show that, despite the differences among the algorithms, they produce remarkably similar task trees. This may be partially explained by our dataset, the universal FOON. For instance, there is only one quesadilla recipe in the universal FOON, so each algorithm goes down the same path when searching for that recipe. In general, using the heuristic functions increases the number of units in the resulting task trees. This is no surprise because the goal of the heuristics we chose is safety, not speed. BFS and IDS are only concerned with finding the goal as quickly as possible, while the heuristics add additional constraints to the search. To measure the performance of heuristic (1), we calculated the average success rate of the motions used in the task trees created by BFS and GBFS.

Table 3: Average motion success rates for task trees

| Goal node | BFS | GBFS, heuristic (1) |
|---|---|---|
| whipped cream | 0.73 | 0.77 |
| Greek salad | 0.56 | 0.55 |
| macaroni | 0.53 | 0.66 |
| sweet potato | 0.35 | 0.35 |
| ice | 0.6 | 0.8 |
| quesadilla | 0.57 | 0.57 |
| sandwich | 0.62 | 0.62 |

In general, using the heuristic improved the average motion success rates in the generated task trees. For the macaroni and whipped cream task trees, this did not increase the number of functional units in tree.

These experiments show that each algorithm is suitable for different use cases. BFS works well for simple recipes that do not require searching deep within the universal FOON. IDS saves memory and finds longer paths faster than BFS. The searches guided by the heuristics produce task trees that are less risky and more likely be completed by the robotic agent successfully. The choice of which algorithm to use ultimately depends upon which of these factors is most important for the given use case. There is no doubt that each algorithm has a place in robotic motion planning.